\documentclass[10pt,twocolumn,twoside,letterpaper]{IEEEtran}

\usepackage{geometry}
\makeatletter
\def\ps@IEEEtitlepagestyle{
  \def\@oddfoot{\mycopyrightnotice}
  \def\@evenfoot{}
}
\def\mycopyrightnotice{
  {\footnotesize
  \begin{minipage}{\textwidth}
  \centering
  978-1-7281-7693-2/20/\$31.00 \copyright2020 IEEE
  \end{minipage}
  }
}

\usepackage{cite}
\usepackage{amsmath,amssymb,amsfonts}
\usepackage{algorithmic}
\usepackage{graphicx}
\usepackage{textcomp}
\usepackage{subfigure}
\usepackage{xcolor}
\usepackage{nopageno}
\usepackage{array}

\usepackage{url}

\begin{document}

\title{Sinogram Denoise Based on Generative Adversarial Networks}

\author{Charalambos Chrysostomou
\thanks{C. Chrysostomou is with the computation-based Science and Technology Research Center, The Cyprus Institute, 20 Konstantinou Kavafi Street, 2121, Aglantzia, Nicosia, Cyprus (e-mail: c.chrysostomou@cyi.ac.cy)}

}

\maketitle
\let\thefootnote\relax\footnote{Manuscript received December 20, 2020}

\begin{abstract}
A novel method for sinogram denoise based on Generative Adversarial Networks (GANs) in the field of SPECT imaging is presented. Projection data from software phantoms were used to train the proposed model. For evaluation of the efficacy of the method Shepp Logan based phantom, with various noise levels added where used. The resulting denoised sinograms are reconstructed using Ordered Subset Expectation Maximization (OSEM) and compared to the reconstructions of the original noised sinograms. As the results show, the proposed method significantly denoise the sinograms and significantly improves the reconstructions. Finally, to demonstrate the efficacy and capability of the proposed method results from real-world DAT-SPECT sinograms are presented.  

\end{abstract}


%
\IEEEpeerreviewmaketitle

\section{Introduction}

Computed tomography (CT) is one of the most broadly used imaging modalities, and its usage has been consistently growing over the past decades \cite{de2009projected} with Positron Emission Tomography (PET) \cite{cherry2001fundamentals, vaquero2015positron}, and Single Photon Emission Computerized Tomography (SPECT) \cite{wernick2004emission, madsen2007recent, mariani2010review}. Even though computed tomography is an essential intermediary in medicine, one major disadvantage is the type of radiation used, which can be dangerous to patient health. In order to reduce the radiation dosage in safe levels, the number of projections or the radiation intensities needs to be limited, that produces less desirable results. Reconstructing a high-quality CT image from such measurement is a great challenge. Current reconstructions methods under-perform when a low number of projections is available or when the measurements are low quality and noisy. In this paper, we proposed a new methodology based on deep learning methodologies to denoise sinograms and improve the reconstruction accuracy of existing methodologies while keeping the radiation dose low.

The paper is organised as follows: Section \ref{sec:data} presents the data generated and used for training the proposed model, Section \ref{sec:cnn}, presents the proposed model. Section \ref{sec:results}, presents the results and discussions finally Section \ref{sec:conclussion} is conclusions.

\section{Methods and Materials}

\subsection{Training Data}
\label{sec:data}

\begin{figure}[htp]
\centering
        \includegraphics[width=9cm]{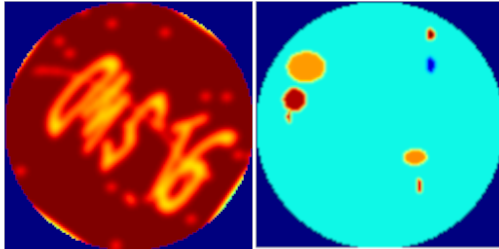}
\caption{Sample software phantoms randomly generated in order to train the proposed model}
\label{fig:random}
\end{figure}

For the training of the proposed method, 200,000 software phantoms were randomly generated. For each phantom, the image of the "true" activity distribution sampled on a rectangular grid of $128 \times 128$ pixels size. Sets of vectorised projections (sinograms) were generated from the "true" space by simulating 32 projections, evenly spaced in 360 degrees. The generated projections obtained from the phantom images were further randomised with a Poisson probability distribution to provide the noisy sets of projections. Samples of the randomly generated phantoms are shown in figure \ref{fig:random}.

\subsection{Proposed Model}
\label{sec:cnn}

\begin{figure*}[!htp]
\centering
\includegraphics[width=16cm]{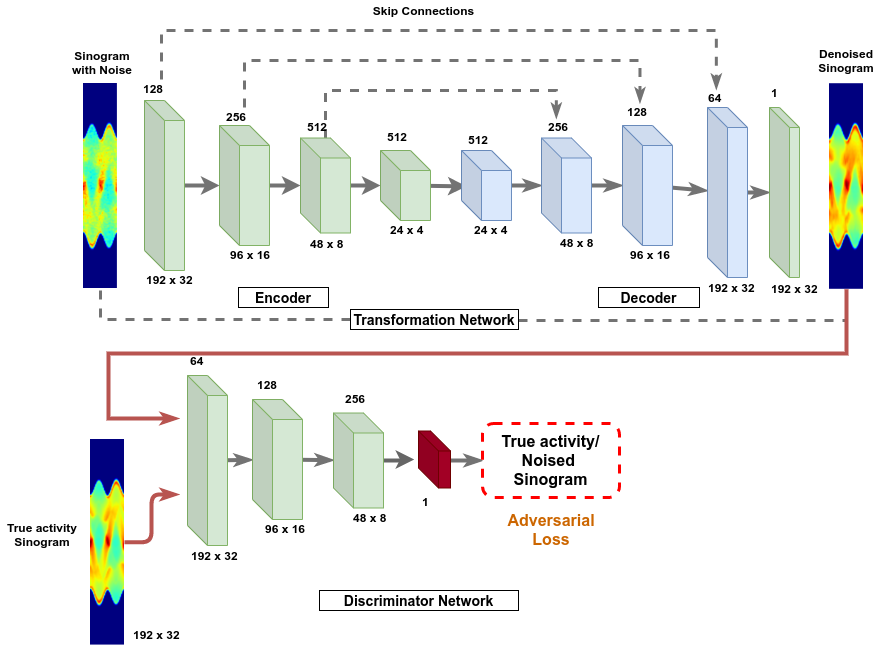}
  \caption{Structure of the proposed model based on conditional generative adversarial networks}
\label{fig:proposed_model}
\end{figure*}

Deep learning methodologies have already successfully applied in image analysis, and classification \cite{chrysostomou2018reconstruction, chrysostomou2019spect}. The proposed novel methodology is based on conditional Generative Adversarial Network (cGAN) \cite{isola2017image}. The GAN method is an approach for training a deep convolutional neural network for image-to-image translation tasks. A generative adversarial network (GAN) is a type of machine learning methodology \cite{goodfellow2014generative} where two neural networks, the generator and the discriminator, challenge each other in a zero-sum style game in which each network's improvements or losses are precisely balanced with the improvements and losses of the other network. The generator learns to create credible data while the discriminator learns to recognise the generator's artificial generated data from the original data. GANS, given a dataset, can generate new data with the same statistics as the training set. A GAN model trained on images with specific characteristics can produce new images that have common characteristics with the original images, and look authentic to external observers. Originally, GANs were proposed as generative models for unsupervised learning but have demonstrated to be valuable in supervised learning \cite{isola2017image}, semi-supervised learning \cite{salimans2016improved} and reinforcement learning \cite{ho2016generative}. The purpose of this work is to denoise sets of sinograms that can be used to reconstruct as accurately as possible the original activity using SPECT. GAN models can create new random credible samples of denoised sinograms from a given dataset. However, a method is needed to control and constrain the characteristics of sinograms that are generated while attempting to calculate the connection between latent space input to the generated images.

The proposed method, as shown in Figure \ref{fig:proposed_model} is based on conditional generative adversarial networks (cGAN) \cite{isola2017image}, where a given input image is used as the condition and constrain for the generated image, using a noised and denoised dataset of sinograms. The proposed model changes the loss function so that the generated denoised sinogram is a probable translation of the input noised sinogram as well as in the content of the target domain. 

The proposed architecture is separated into two parts, the transformation network and the discriminator network. The transformation network consists of the encoder and decoder subparts. The encoder consists of four convolutional layers of 3x3 kernel and utilising the rectified linear activation function (ReLU) \cite{agarap2018deep}. For each layer, the number of kernels increases, beginning with 128 kernels for the first block, 256, 512 and 512 for the second, third and fourth blocks respectively. The first three convolutional layers of the encoder are followed by a 2x2 max pooling layer \cite{goodfellow2016deep}. The decoder consists of four convolutional layers of 3x3 kernel and ReLU as the activation function. The first three convolutional layers are followed by a 2x2 up-sampling layer \cite{dumoulin2016guide}. For each layer, the number of kernels decreases, beginning with 512 kernels for the first layer, 256, 128 and 64 for the second, third and fourth layer respectively. The layers of the encoder and decoder of the transformation network are connected with skip connections by applying a concatenation operator as Figure \ref{fig:proposed_model} shows. Finally, the output layer is a convolutional layer of 3x3 kernel and linear activation function. 

The discriminator network consists of three convolutional layers of 3x3 kernel and ReLU as the activation function. For each layer, the number of kernels increases, beginning with 64 kernels for the first layer, 128 and 256 for the second and third layers respectively. The last layer of the discriminator is a fully connected layer of one neuron and sigmoid activation function. 

\section{Results and Discussions}
\label{sec:results}

\begin{figure}[!htp]
\centering
        \includegraphics[width=9cm]{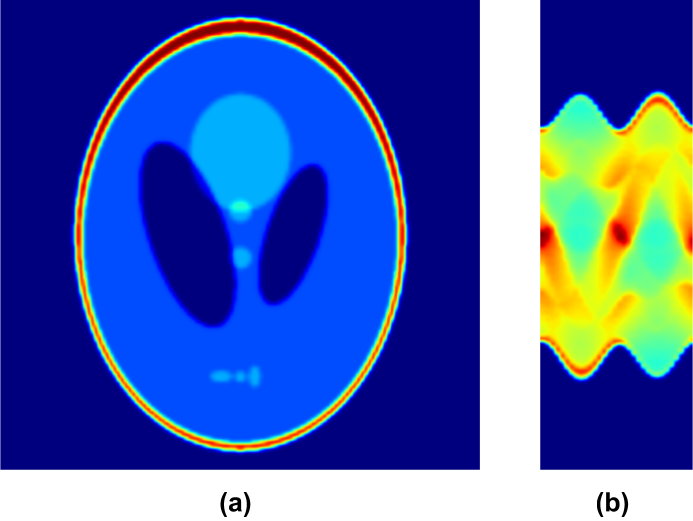}
\caption{(a) Shepp Logan Phantom used to evaluate and demonstrate the capabilities of the proposed method (b) Sinogram generated from the Shepp Logan Phantom without the addition of noise}
\label{fig:original_phantom_sinogram}

\end{figure}
\begin{figure*}[htp]
\centering
\includegraphics[width=17cm]{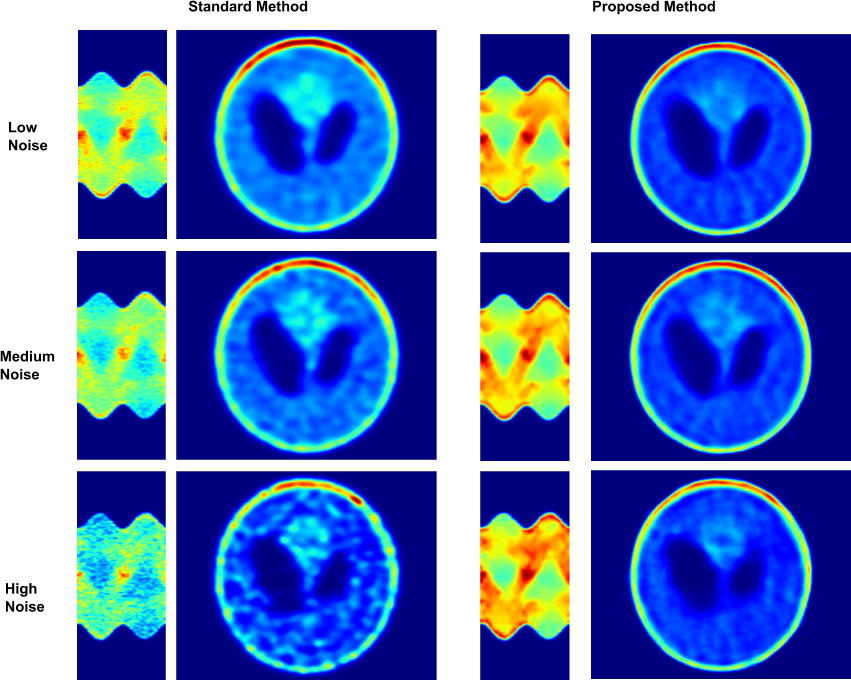}
  \caption{Evaluation and comparison of standard OSEM versus the proposed method by using the Shepp Logan Phantom.}
\label{fig:results}
\end{figure*}
\begin{figure*}[htp]
\centering
\includegraphics[width=17cm]{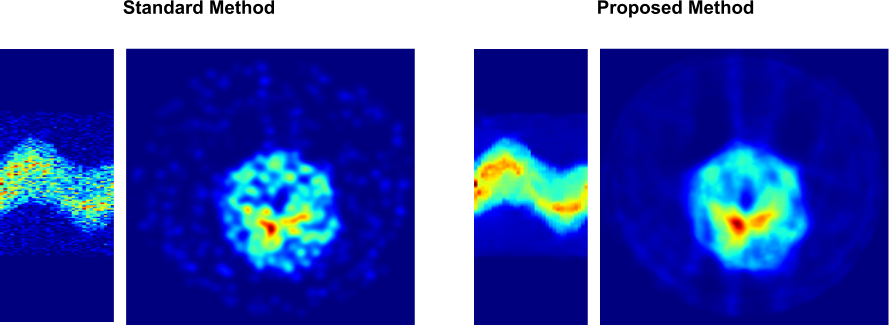}
  \caption{Results of the real-world DAT-SPECT sinogram reconstructions by using OSEM for the proposed method versus the standard method}
\label{fig:real_results}
\end{figure*}

The efficiency of the proposed method and comparison against existing methods, the  Mean Absolute Percentage Error (MAPE), Mean Square Error (MSE), Structural Similarity (SSIM) Index \cite{wang2004image}, and the Peak signal-to-noise ratio (PSNR) were used.  The Shepp Logan phantom \cite{shepp1974fourier} (Figure \ref{fig:original_phantom_sinogram} (a)), along with the original sinogram generated without the addition of noise (Figure \ref{fig:original_phantom_sinogram} (b)) was used to evaluate and demonstrate the capabilities of the proposed method. Table \ref{denoise_results_table} shows the results of the denoised method where three different levels of noise where added, low, medium and high, versus the original noise-free sinogram as presented in Figure \ref{fig:original_phantom_sinogram} (b). Table \ref{osem_results_table} and Figure \ref{fig:results} and shows the reconstruction results based on standard OSEM versus the proposed method. As the results show, the proposed method is capable to denoise the sinograms in multiple levels of noise and significantly improves the reconstructions. Finally, in order to test the efficacy of the proposed model, we used real-world DAT-SPECT sinograms and produced reconstructions based on the standard OSEM and proposed method, as showed in Figure \ref{fig:real_results}. As the results demonstrate, the proposed methodology is capable of being trained using software phantoms, which are computationally inexpensive to generate and been applied to real-world scenarios where the generation or collection of data is limited.

\begin{table}[ht]
\renewcommand{\arraystretch}{1.5}
\centering
\caption{Sinogram Denoising Results}
\label{denoise_results_table}

\begin{tabular}{lllll}
\hline

\textbf{Noise} & \textbf{MAPE} & \textbf{MSE} & \textbf{SSIM} & \textbf{PSNR} \\
\hline

Low            & 4.80\%  & 0.0009       & 0.975         & 30.48         \\
Medium         & 5.83\%  & 0.0010       & 0.971         & 29.91         \\
High           & 8.39\%  & 0.0054       & 0.930         & 22.65        \\
\hline

\end{tabular}
\end{table}

\begin{table}[ht]
\renewcommand{\arraystretch}{1.5}
\centering
\caption{Reconstruction Results based on OSEM}
\label{osem_results_table}

\begin{tabular}{lllllll}
\hline
& \multicolumn{3}{c}{\textbf{Standard Method}} & \multicolumn{3}{c}{\textbf{Proposed Method}} \\
\hline
\textbf{Noise Level} & \textbf{MSE}    & \textbf{SSIM}    & \textbf{PSNR}   & \textbf{MSE} & \textbf{SSIM} & \textbf{PSNR} \\
Low                  & 0.0082          & 0.82             & 20.89           & 0.0040       & 0.89          & 23.94         \\
Medium               & 0.0087          & 0.80             & 20.58           & 0.0045       & 0.89          & 23.48         \\
High                 & 0.0099          & 0.75             & 20.05           & 0.0061       & 0.84          & 22.17\\
\hline
\end{tabular}

\end{table}

\section{Conclusions}
\label{sec:conclussion}

In this study, we demonstrate the capability of GANs to perform sinogram denoise in SPECT imaging. As the results show, the proposed method significantly improves the results and outperforms the standard OSEM method. Although the phantom and the results presented in this paper is appropriate for illustrating the capabilities of the proposed method, further experimentation is needed for evaluating the potential application of the method in clinical studies.

\bibliography{ref} 
\bibliographystyle{IEEEtran}

\end{document}